\documentclass{article}

\usepackage[preprint]{neurips_2023}

\usepackage[utf8]{inputenc} 
\usepackage[T1]{fontenc}    
\usepackage{hyperref}       
\usepackage{url}            
\usepackage{booktabs}       
\usepackage{amsfonts}       
\usepackage{nicefrac}       
\usepackage{microtype}      
\usepackage{xcolor}         
\usepackage{natbib}
\usepackage{graphicx}
\usepackage{amsmath}
\usepackage{multirow}
\usepackage{enumitem}
\DeclareUnicodeCharacter{2212}{-}

\title{Multi-Agent Learning of Efficient Fulfilment and Routing Strategies in E-Commerce}

\author{%
  Omkar Shelke \\
  TCS Research\\
  Mumbai, India \\
  \texttt{shelke.omkar@tcs.com} \\
  \And
  Pranavi Pathakota \\
  TCS Research\\
  Mumbai, India \\
  \texttt{p.pranavi@tcs.com} \\
  \And 
  Anandsingh Chauhan \\
  TCS Research\\
  Mumbai, India \\
  \texttt{anandsingh.chauhan@tcs.com} \\
  \And 
  Harshad Khadilkar \\
  TCS Research, IIT Bombay\\
  Mumbai, India \\
  \texttt{harshad.khadilkar@tcs.com} \\
  \texttt{harshadk@iitb.ac.in} \\
  \And 
  Hardik Meisheri \\
  TCS Research\\
  Mumbai, India \\
  \texttt{hardik.meisheri@gmail.com} \\
  \And 
  Balaraman Ravindran \\
  RBCDSAI, IIT Madras\\
  India \\
  \texttt{ravi@cse.iitm.ac.in} \\
}

\begin{document}

\maketitle

\begin{abstract}

This paper presents an integrated algorithmic framework for minimising product delivery costs in e-commerce (known as the cost-to-serve or C2S). One of the major challenges in e-commerce is the large volume of spatio-temporally diverse orders from multiple customers, each of which has to be fulfilled from one of several warehouses using a fleet of vehicles. This results in two levels of decision-making: (i) selection of a fulfillment node for each order (including the option of deferral to a future time), and then (ii) routing of vehicles (each of which can carry multiple orders originating from the same warehouse). We propose an approach that combines graph neural networks and reinforcement learning to train the node selection and vehicle routing agents. We include real-world constraints such as warehouse inventory capacity, vehicle characteristics such as travel times, service times, carrying capacity, and customer constraints including time windows for delivery. The complexity of this problem arises from the fact that outcomes (rewards) are driven both by the fulfillment node mapping as well as the routing algorithms, and are spatio-temporally distributed. Our experiments show that this algorithmic pipeline outperforms pure heuristic policies.


\end{abstract}

\section{Introduction}

\subsection{Motivation}

Efficient operation of supply chains has been a problem area of interest for several decades \citep{holt1955linear,haley1973inventory,lambert2000issues,Path23}. However, it is still an open problem because of the constant disruptions in this field. In recent years, the advent of electronic commerce (abbreviated to `e-commerce') has created new challenges. Specifically, the traditional retail supply chain network consisting of warehouses and stores has been abruptly extended to include individual customer delivery locations. The fulfillment of e-commerce orders from a warehouse (`fulfillment node') requires two types of decisions; first, the selection of the fulfillment node from which the order is to be served, and second, the routing of vehicles from these nodes to a set of delivery locations. We define the total cost resulting from these decisions as the cost-to-serve, or C2S. Presently, these decisions are based on simple heuristics or rules -- the nearest fulfillment node to every customer location is chosen, and then vehicles are sent to individual clusters of delivery locations. Our hypothesis in this paper is that the fulfillment node selection and vehicle routing are interdependent problems, and should be solved using an integrated approach. This improves the efficiency of the system both from resource utilization as well as turn around time. We show that a reinforcement learning algorithm trained to perform this task is able to outperform heuristics on several key performance indicators, and has the ability to adapt to changes in business objectives using weights on the individual reward terms.

\subsection{Overview}

\begin{figure}
    \centering
    \includegraphics[width=0.55\columnwidth]{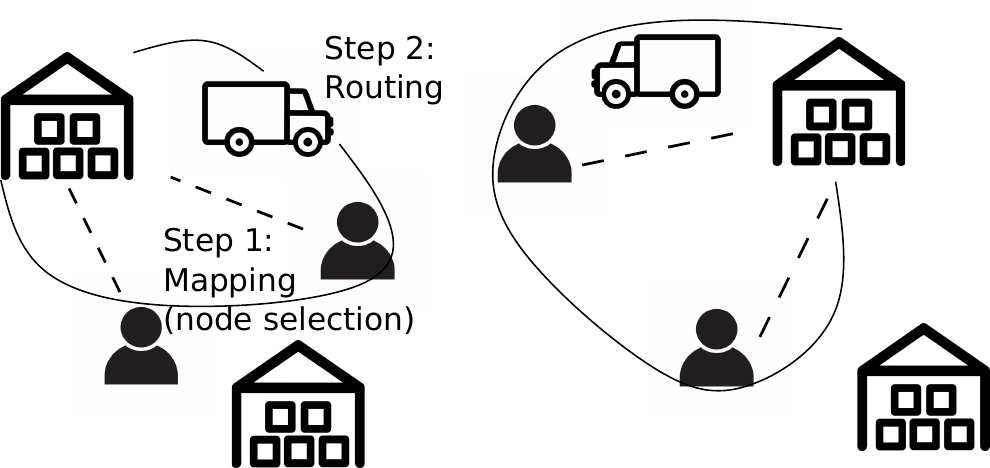}
    \caption{Illustration of e-commerce supply chain. The proposed algorithm learns to optimise the node selection and routing steps in an integrated fashion.}
	\label{fig:Supply-chain}
	\vspace{-1.5em}
\end{figure}

We consider the scenario of last-mile delivery, where the goal of the retail e-commerce business is to serve dynamic customer orders with minimum operating cost. The portion of interest in the supply chain is shown by the schematic in Figure \ref{fig:Supply-chain}. An analysis of existing work in this area is given in Section \ref{sec:literature}. In this work, we assume that there is a single type of product which can be ordered in arbitrary quantity by every customer. The customers `pop up' via a stochastic process in random locations and with demand drawn from a probability distribution. There is a minimum and maximum time corresponding to each order, within which delivery must happen. The warehouse locations are fixed, and the inventory of each warehouse is replenished using an external process (outside the scope of this work). The task of the fulfillment algorithm is to pick the warehouse for fulfilling each order, or to choose to defer the fulfillment to a future time. All orders being fulfilled in the present time step are then serviced by a vehicle routing agent, which takes into account vehicle capacity, travel times, and customer time windows. As described in Section \ref{sec:problem}, we consider multiple performance indicators, including proportion of successful deliveries, distance travelled, and vehicle capacity utilisation.


The resulting problem falls under the class of combinatorial optimisation (CO). While exact optimisation methods such as Mixed-Integer Linear Programming (MILP) perform well for static instances with fixed locations, they are less effective in the e-commerce context where situations are constantly changing. In Section \ref{sec:method}, we use graph neural networks (GNN) to model the graph of customer and warehouse locations. GNNs can adapt to dynamic changes in nodes and can be combined with other learning techniques for downstream applications. We have found that by using the learned embeddings from graph training, a reinforcement learning (RL) policies are able to effectively assign fulfillment nodes and scheduling vehicle for routing. Section \ref{sec:results} shows that the resulting algorithm outperforms state-of-the-art heuristics on the various performance indicators. The contributions of this paper are as follows.

\begin{enumerate}[topsep=0pt, itemsep=0pt, leftmargin=15pt]
    \item We formalise the fulfillment node selection and vehicle routing scenario as an integrated optimisation problem.
    \item We propose an algorithm that uses GNN for modeling supply chain entities and RL based policies for the downstream tasks of fulfilment and vehicle routing.
    \item We show that the proposed approach is able to outperform separately defined heuristics for fulfillment node selection and vehicle routing on a number of performance indicators, and for a variety of customer demand distributions.
\end{enumerate}

\section{Literature Review} \label{sec:literature}

This is an initial approach of integrating Cost-to-Serve (C2S) with routing of vehicles to a set of delivery locations. There is limited literature available on C2S, which usually adopts a process management perspective \citep{rwblog}. The term {C2S} involves recognizing that the cost drivers in a supply chain can vary for different products and sales channels  \citep{cooper97, alan98, robert01}. In recent work, C2S was defined as a sequential decision-making problem and proposed a learning based approach minimise it \citep{Path23}. However, the study did not include the vehicle routing problem (VRP) in the decision loop. VRP by itself is a well-known NP-hard problem in the area of combinatorial optimisation \citep{article, article_vrp}. The objective is to determine the optimal routes for one or more vehicles travelling to a specified set of customer locations. The problem is equivalent to the travelling salesman problem (TSP); if there is only one vehicle with no restrictions on its capacity, fuel, or range. 


Capacitated VRP with time windows (CVRP-TW) is a constrained version of VRP that imposes a limit on the load carried by each vehicle, and the time required for each leg and service on the journey. It has been solved using the following approaches:  heuristic or rule-based approach, linear programming \citep{linear_vrp, 10.1016/j.cor.2006.11.006}, randomised search or meta-heuristics approach such as genetic algorithm \citep{BAKER2003787}, neighbourhood search, ant colony optimisation based approach \citep{BELL200441} and more recent learning-based approach such as  attention mechanism and reinforcement learning \citep{nazari2018reinforcement, gupta2022deep} or a combination of reinforcement learning and satisfiability solver \citep{khadilkar2022solving}. The heuristics approaches have performed excellent for the dynamic VRP problem, and are known to produce close to optimal results in the literature, the only drawback is to design a fix set of rules for specific instances while the learning based approaches can adapt to new instances quickly.

Recently, studies have recognised the importance of neighbourhood information in solving practical combinatorial optimisation problems. A popular approach is using Graph Neural Networks (GNNs). These methods have shown prominent results in wide range of applications ranging from recommender systems \citep{wu2022graph}, traffic analysis \citep{wang2018spatio}, drug discovery \citep{gaudelet2021utilizing} to resource allocation \citep{LIU20221211}. An end-to-end RL training with GNN is proposed in \citep{khalil2017learning}, where a learning based greedy algorithm is learnt for combinatorial problems by training GNN and a Deep Q-Network (DQN) \citep{mnih2015human} together. Their main focus is on learning an approximate greedy policy which can adapt to slight change in data distribution from training by adapting GNNs. In this paper, we focus on using GNN techniques to aid the learning agent by producing better embeddings over unseen graphs of customer demand.

The proposed learning based framework aims to provide an integrated solution to the single-product, multi-depot C2S, including VRP with the time window and capacity constraints. The problem formulation proposed in Sec. \ref{sec:problem}. The methodology is described in Sec. \ref{sec:method}, including the graph embedding generation, training of RL agent for C2S, and the VRP heuristic used in this paper. Experiments and results are described in Sec. \ref{sec:results}.

\section{Problem Formulation} \label{sec:problem}

\subsection{High-level description}

The system described is a multi-objective optimization problem that aims to fulfill customer orders while minimizing the total cost incurred by the enterprise. The entities considered in this system are warehouses, vehicles, and customers. The cost of fulfilling an order is determined by several factors, including the distance traveled between various nodes, the capacity utilisation  of the vehicles, and the number of customers served. The decision variables are the allocation of warehouses to customer orders, and the routes generated for the vehicles that deliver the orders. The characteristics are,

\begin{itemize}[topsep=0pt, itemsep=0pt, leftmargin=8pt]
    \item At each time step, multiple customer orders may arise, each specifying a desired product quantity and a delivery time window. Failure to meet this window incurs a penalty. 
    \item A reinforcement learning approach is employed to calculate vehicle routes, accounting for factors like vehicle capacity, travel times for each leg, and customer service times.
    \item Split deliveries is not allowed; entire product quantity must be served by allotted warehouse.
    \item Each vehicle adheres to a uniform maximum capacity constraint, consistent across all vehicles, and must return to its starting depot after serving customers on its route.
    \item Consistent speed across all vehicles reuslts in travel times directly corresponding to distance.
    \item While each warehouse can dispatch an unlimited number of vehicles, the objective is to minimize the number of vehicles and the number of trips. Note that the proposed algorithm generalises to finite number of vehicles without change.
    \item Warehouses are replenished to their maximum capacity at fixed intervals, regardless of demand.
\end{itemize}
 
As orders are fulfilled, inventory is depleted from the chosen warehouse(s), leaving less available for future orders. The C2S policy must consider the geographical distribution of customers, balancing the overall end-to-end cost, while generating real-time optimal node fulfillment taking into account dynamic demand, location and time windows of the customers.

\subsection{Mathematical formulation}

In the description below, we formally specify the notation, constraints, and objectives of the problem.

\textbf{Warehouse} : We assume the existence of $N$ warehouses indexed by $o=\{1,2,\ldots,N\}$, and denote the $o^\mathrm{th}$ warehouse location by $w_o = (w_{o,x},w_{o,y})$. In this paper, we define a 2D grid on the range $[-100,100]\times [-100,100]$ and fix the location of 4 warehouses, one in the center of each quadrant of the grid. The warehouses stock a single type of product and their maximum inventory levels are identical. Each warehouse is restocked to its maximum level $P_{o}^\mathrm{(max)}$ after every $\frac{T}{2}$ time steps. 

\textbf{Customer demand generation}: 
Each episode consists of $\tau$ time units. After every $T<\tau$ units\footnote{In this paper, we use $T=100$ and $\tau=1000$.}, two sets of events occur. First, the warehouse inventory levels are set to their maximum levels (as described above). Second, a new set of customer demands is generated as follows. 

\begin{itemize}[topsep=0pt, itemsep=0pt, leftmargin=8pt]
    \item At every interval $T$, the number of unique customers is first generated from a uniform distribution ranging between 200 to 300 customers, which results in approximately 2500 customers over $\tau$ time units in an episode.
    \item  The integer quantity demanded by the customer is chosen uniformly in the range $U[1,10]$.
    \item If the current time is $t$, the minimum and maximum time window limits are generated using the formulae,
    \begin{equation*}
    [\theta_{i,\min}, \theta_{i,\max}] = [t + U(0.2\;T, 0.8\;T), \theta_{i,\min} + U(0.1\;T, 2\;T)]
    \end{equation*}
where $U(a,b)$ denotes the uniform distribution over the range $(a,b)$.
\end{itemize}

We denote customer $i$ by the symbol $c_i$, and their characteristics are defined by the n-tuple $(m_i,x_i,y_i,t_i,\theta_{i,\min},\theta_{i,\max})$ where $m_i$ is the demand quantity of product, $(x_i,y_i)$ is the location in a 2D grid, and $t_i$ is the time step at which the customer is generated. Each customer has a stipulated time window  $[\theta_{i,\min},  \theta_{i,\max}]$ during which order must be delivered.

Based on the problem specifications, we derive some important quantities from the point of view of the planning process. The Euclidean distance from  warehouse $o$ to customer $i$ is denoted by $d_{o,i}$, and distance between two customers $i,j$ by $d_{i,j}$. As soon as new customer demands are generated, the C2S agent is required to compute allocation decisions for all open demands (newly generated customers as well as ones deferred from previous time steps). The decisions consist of either allocation to a specific warehouse, or a deferral of the customer to the next time step.

\textbf{Vehicle} : 
We assume that vehicles can be spawned as required from each depot. Once a vehicle is spawned for the first time, it remains available for reuse for the rest of the episode. The vehicle characteristics are homogeneous, with maximum capacity Q and average speed $v$. We ensure that the demand generated for a single customer $m_i$ cannot exceed the vehicle capacity $Q$, thus ensuring that multiple vehicles are not required for serving individual customers. The VRP agent described in Section \ref{sec:method} is used to compute the routes according to the decisions made by the C2S agent. Note that if a vehicle completes its assigned tour within $T$ time units, it will be available to serve customers generated in the next set of demand generation. However, this is not a requirement imposed by the environment (trip may take longer than $T$ time units).

Our vehicle routing approach considers constraints related to both capacity and time windows, resulting in the capacitated vehicle routing problem with time windows (CVRP-TW); and it assumes that customer-warehouse allocations are known at each time step. In our framework, the VRP agent described in section assumes that at timestep t, the C2S agent provides information about customer-warehouse pairs, where each customer's location is represented as $c_{i}^{(t)} = (c_{i,x}^{(t)}, c_{i,y}^{(t)})$, with a demanded load of {$m_{i}^{t}$} and service time windows $[T_{i,min}, T_{i,max}]$, $\in \mathbb{R^{+}}$, with T$_{i,min}$ < T$_{i,max}$, and {where i $\in$ C}. Then the objective of the problem \citep{sultana2021fast} is to find the total distance $J$ that minimises,




\begin{equation}
    J = \min_{a_{*},\;f_{*},\;l_{*}} \left({\sum_{i,j,k} d_{i,j}\,a_{i,j,k}} + \sum_{i,k} d_{o,i}\,f_{i,k} + \sum_{i,k} d_{o,i}\,l_{i,k} \right),
    \label{eq:objective}
\end{equation}
where $d_{i,j}$ is the distance from customer $i$ to customer $j$, $d_{o,i}$ is the distance from origin (depot) to customer $i$, $a_{i,j,k}$ is an indicator variable which is 1 if vehicle $k$ goes directly from customer $i$ to customer $j$, $f_{i,k}$ is an indicator variable which is 1 if customer $i$ is the first customer served by vehicle $k$, and $l_{i,k}$ is a similar indicator variable which is 1 if customer $i$ is the last customer visited by vehicle $k$. Apart from constraints on $a_{i,j,k}$, $f_{i,k}$, and $l_{i,k}$ to take values from $\{0,1\}$, the other constraints are defined below in brief. A detailed description can be found in \citep{sultana2021fast}.

Every customer must be served within its time window. If vehicle $k$ visits $i$ at time $t_{i,k}$ , then 
\begin{equation}
T_{i,\min} \leq t_{i,k} \leq T_{i,\max}, \text{ if }f_{i,k}=1\text{ or }\exists j\text{ s.t. }a_{j,i,k}=1 \label{eq:tw}
\end{equation}
The total load served by a vehicle must be at most equal to its capacity $Q$. This is formalised as,
\begin{equation}
\sum_{i,j} m_{j}\,a_{i,j,k} + \sum_i m_{i}\,f_{i,k} \leq Q \quad \forall k \label{eq:maxload}
\end{equation}
Travel time constraints are applicable between any two locations, based on the distance between them, the speed $v$ at which vehicles can travel, and the fixed service time $\Delta$ required at each customer.
%
    %

\begin{minipage}{.45\linewidth}
\begin{equation*}
  t_{i,k} \geq \frac{d_{o,i}}{v} \text{ if }f_{i,k}=1  
\end{equation*}
\end{minipage}%
and 
\begin{minipage}{.5\linewidth}
\begin{equation}
  t_{i,k} \geq t_{j,k} + \Delta + \frac{d_{j,i}}{v} \text{ if } a_{j,i,k}=1 \label{eq:intercustomer} 
\end{equation}
\end{minipage}
%








The vehicle capacity utilisation  ratio for this trip is given by, $\frac{\sum_{i,j} m_{j}\,a_{i,j,k} + \sum_i m_{i} \,f_{i,k}}{Q}$. 
%
%
On each leg of the route, a minimum travel time proportional to the distance is enforced. The service for the first customer on the route can start at $(d_{o,i}\,f_{i,k})/v$ or later. A fixed service time of $\Delta$ units is assumed, which means the vehicle must stay at the customer location for this time. The service for the next customer can then start after a further ($d_{i,j}\,a_{i,j,k})/v$ time units, and so on.

\begin{figure}
    \centering
    \includegraphics[width=0.75\columnwidth]{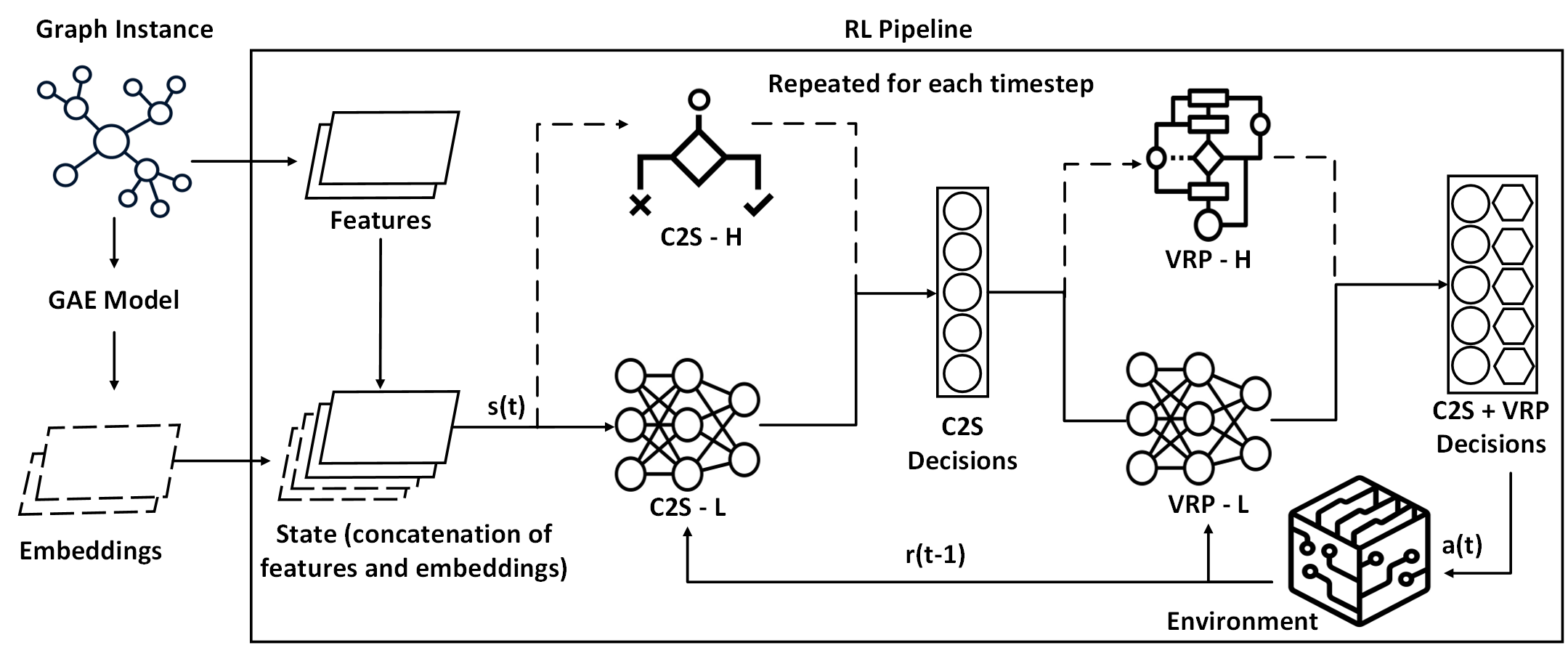}
    \caption{Proposed RL Framework}
	\label{fig:rl-pipeline}
	\vspace{-1.5em}
\end{figure}

\section{Methodology} \label{sec:method}

The proposed solution composed of three phases (i) Training a Graph Auto Encoder \citep{zhang2022graph} to get the node embeddings of the graph, and (ii) using these learnt embeddings along with other state features to train a DQN \citep{mnih2015human} agent to assign warehouses to customers. The third phase incorporates a VRP agent to compute routes and assign vehicles. 

\subsection{Generating graph embeddings} 

We train a graph auto encoder \citep{zhang2022graph} using the \texttt{Pytorch} \texttt{geometric} library \citep{Fey/Lenssen/2019} in Python, to generate node embeddings for each customer. The GAE requires the input of a graph $G = (V_{t}, E_{t})$. In our cases, the vertex set $V_t$ is composed purely of the customer locations as features. The edges (and hence adjacency matrix) is constructed such that there is an edge between two customers if distance between them falls with in nearest distance warehouse neighbourhood radius. 
%
%
The Encoder of GAE model takes input features such as the customer locations and adjacency matrix, and learns an embedding for each node of the graph in a 2-dimensional embedding space. The decoder, which is used to reconstruct the graph, is based on a Euclidean distance metric that calculates the distance between two embeddings. Specifically, we define the similarity between two embeddings $e_1$ and $e_2$ by,
\begin{equation*}
\text{similarity}=1-\frac{||e_1-e_2||_2}{\max ||e_1-e_2||_2}.
\end{equation*}


\subsection{C2S Learning Agent (C2S-L) - Fulfillment node selection} \label{subsec:c2s-l}

		
		
		
		
		

C2S problem as defined in previous section can be modelled as a Markov Decision Process ($\mathcal{S}, \mathcal{A}, \mathcal{T}, \mathcal{R}, \gamma$), where, $\mathcal{S}$ denotes the current state of the system, $\mathcal{A}$ represents action space (mapping customer $c_i$ to a warehouse $j$), $\mathcal{T}:(\mathcal{S}, \mathcal{A}) \rightarrow \mathcal{{S}}$ represents transition probabilities, $\mathcal{R}$ is the set of rewards, and $\gamma$ is the discount factor for future rewards. In prior work \citep{Path23}, the task of C2S was only to allocate a fulfillment node (warehouse) to each customer, optimizing a multi-objective function consisting of various costs. They assumed that delivery of order happens instantaneously as soon as the decision was made by C2S, excluding the routing for vehicles and also ignored time window constraints as a result. 

The graph embeddings produced by the GAE, are concatenated with other state features (Table \ref{tab:c2s_statespace} mentioned in Appendix) to result in an input vector of size 19 for each customer. The action space for each customer is of size $N+1=5$, where $N$ is the number of warehouses. The first $N$ actions correspond to assignment to that particular warehouse ID, while the last action is a choice to defer the customer fulfillment to the next decision point (after $T$ time units). Proceeding in a first-come-first-served (FCFS) order, the RL agent computes a decision for each customer. This list includes any customers carried over from previous time steps, as well as ones generated in the current time step. Once all the decisions are completed, the customers assigned to each warehouse are communicated to the VRP agent (Section \ref{subsec:vrp-l}) for delivery scheduling and routing. 

\subsubsection{Reward for C2S-L Agent} \label{subsec:reward_c2s-l}

The step reward for the RL agent is composed of the following components:
\begin{itemize}[topsep=0pt, itemsep=0pt, leftmargin=8pt]
    \item Transportation reward, composed of two components. First, we assign a negative reward $D_i$ proportional to the straight-line distance from the warehouse to the location of customer $c_i$. The second term is a negative reward $L_i$ proportional to the distance travelled by the vehicle in that particular trip, equally apportioned to all the customers being served in that trip. If the vehicle has a round-trip distance of $Z$ for a trip that serves $r$ customers, then $L_i\propto - Z/r$. $D_i$ is in the range\footnote{In the 2D space, the farthest euclidean distance between a customer and a warehouse is $1.5\sqrt{2} \approx 2.12$ } of $[-2.12, 0]$ , and $L_i$ is the range of $[-1, 0]$.
    \item A fixed fulfillment reward $F_i=1$ when customer $c_i$ is assigned to a warehouse, and $0$ if the customer is deferred to a future time step.
    \item A negative reward $U_i$ proportional to the empty space on the vehicle when it starts on a trip, given equally to all the customers served in that trip. It also falls in the range $[-1,0]$. The overall reward function is defined by,
\end{itemize}
 \begin{equation}
    \text{reward}(c_i) = a_{1} \; ( D_{i} + L_{i} ) +  F_{i} + a_{2} \; U_{i}, 
    \label{eq:reward_c2s}
\end{equation}
where {$a_{1}$ and $a_{2}$} are user-defined constants. Apart from this, a fixed penalty of $-10$ is assigned if a customer is dropped completely (not served before the time window $\theta_{i,\max}$ elapsed). Note that the terms $L_i$ and $U_i$ depend on the route computed by the VRP agent. There is a special case for reward computation when the customer is deferred to a future time step. For this decision, the quantities from (\ref{eq:reward_c2s}) cannot be computed immediately. We therefore use a backward reward computation (in the Monte-Carlo style) from the time step when the customer is actually served. We define the reward as,
\begin{equation}
    \text{reward}(c_i,deferred) = \gamma^{h} \; ( a_{1}( D_{i} + L_{i} ) +  F_{i} + a_{2} U_{i}) 
    \label{eq:reward2}
\end{equation}
where h denotes the number of times service for deferred for $c_i$, and the remaining quantities correspond to when the customer was actually served. The calculated expected return is used as the reward in the TD-error calculation in the Bellman equation, which is used to train the model on samples that have the holding (deferred) action selected.

\subsection{VRP learning Agent (VRP-L) - Routing and Scheduling} \label{subsec:vrp-l}

{The problem of routing the vehicle to serve the customer is a sequential decision making problem and modelled as an MDP. The VRP-L agent is adapted from  \citep{khadilkar2022solving}, where the input to agent includes information corresponding to customer-vehicle pairs post C2S decisions. The agent in \citep{khadilkar2022solving} which is designed for single depot problem is applied to Multi-Depot problem in this study, by a parallel process of decisions at different depots as the customers allocated to particular depot must be fulfilled by a vehicle originating from that depot. We have incorporated official codebase for VRP-L \citep{khadilkar2022solving} in our study and scaled it for multi-depot and further details corresponding to it mentioned in appendix \ref{subsec:arch}.

\subsubsection{Reward for VRP-L Agent} \label{subsec:reward_vrp-l}


In this framework, a vehicle denoted as $k$ caters to a set of $P$ customers indexed by $p$ from a particular warehouse along its route. Each segment of this route has a distance of $d_p$  and necessitates $t_p$ units of time between completing service at the preceding customer and initiating service at the current customer. The range of the neighborhood is determined by $\rho$, and  $\tau$ serves as a time limit, equating to the median travel time between all pairs of customers within the dataset. Consequently, the reward associated with each decision is as follows:


\begin{minipage}{.41\linewidth}
\begin{equation*}
\mathcal{R}_{k,p} = \frac{\rho-d_p}{d_{max}} + \frac{\tau-t_p}{t_{max}} + \gamma^{P-p}\mathcal{R}_{term} 
\end{equation*}
\end{minipage}%
, where
\begin{minipage}{.5\linewidth}
\begin{equation}
\mathcal{R}_{term}=2\,\rho-\frac{1}{P+1}\,\left( \sum_p d_p + D_{return} \right)
\label{eq:terminal}
\end{equation}
\end{minipage}
%


The reward equation is designed to encourage shorter legs in both distance and time. The terminal reward, denoted as $\mathcal{R}_{term}$  as defined in equation (\ref{eq:terminal}) , measures the average distance of individual legs within a vehicle's journey (including the last leg back to the depot) relative to the largest cluster diameter, which is represented as $2\,\rho$.. The objective of training the neural network is to minimize the mean squared error between its output and the actual realized reward $\mathcal{R}_{k,p}$ , for each decision.

\section{Baselines}
\textbf{Fulfillment Node Selection Heuristic (C2S-H)} \label{subsec:c2s_h}:
We refer to this baseline as C2S Heuristic (C2S-H) as it essentially allocates each customer to the nearest warehouse immediately upon their generation. It's worth mentioning that our conversations with experts in the business domain have indicated that this approach closely aligns with the typical practice followed by a majority of retailers.  

\textbf{Vehicle Routing Problem Heuristic (VRP-H)} \label{subsec:vrp_h}: 
The VRP policy is employed to determine the most efficient route for a group of vehicles to deliver customer orders while adhering to constraints like vehicle capacity and customer time-windows. In our current scenario, we employ a straightforward heuristic that relies on the warehouse assignments provided by the C2S agent to fulfill the customer demand. The VRP Heuristic or (VRP-H) is designed to ensure that the vehicles visit and serve the customers within the specified time windows, while also meeting total time and vehicle carrying capacity constraints. While more advanced heuristics are available, we opt for a simple approach with the primary aim at satisfying constraints rather than achieving optimality.  
The procedure for the same is described below which can be carries out in a parallel fashion:


\begin{enumerate}[topsep=0pt, itemsep=0pt, leftmargin=15pt]
    \item Sort the assigned customers at the warehouse according to their time window opening.
    \item  If no vehicle is at the depot, start a new one.
    \item Choose the first feasible customer from the list, based on travel time, time window constraint, and demand quantity. Serve this customer and update time and remaining vehicle capacity.
    \item Repeat the above step until there are no further feasible customers.
    \item Return the vehicle to the warehouse and update its availability time.
    \item Repeat steps 2-5 with additional vehicles until all customers are either (i) served, or (ii) computed to be infeasible for any vehicle. A penalty will be applied in the latter case.
\end{enumerate}

\begin{table}[t]
    \caption{Agent versions with description: values in superscript indicate reference sections.} 
    \label{tab:agents} 
    \begin{tabular}{  c  p{0.75\linewidth} } 
        \toprule
        Agent & Description \\
        \hline
        \textbf{\textit{C2S-H + VRP-H}} & Combination of the C2S-Heuristic and VRP-Heuristic\textsuperscript{\ref{subsec:vrp_h}}, serving as a comprehensive heuristic baseline \\
        \hline
        \textbf{\textit{C2S-L + VRP-H}} & Utilizes the C2S Agent\textsuperscript{\ref{subsec:c2s-l}} along with the VRP-Heuristic\textsuperscript{\ref{subsec:vrp_h}} \\
        \hline
        \textbf{\textit{C2S-H + VRP-L}} & Incorporates the C2S-Heuristic\textsuperscript{\ref{subsec:vrp_h}} and the VRP-Agent defined in\textsuperscript{\ref{subsec:vrp-l}} \\
        \hline
        \textbf{\textit{C2S-L + VRP-L}} & Signifies the use of both agents as learning-based for C2S and VRP. \\
        \hline 
        \textbf{\textit{C2S-P + VRP-L}} & We use a pre-trained model from C2S-L + VRP-H to set the weights for the C2S agent, and the VRP agent also adopts a learning-based approach. Our training strategy unfolds in two phases: first, we exclusively train the C2S Agent with VRP-H, and subsequently, we train the VRP Agent based on decisions made by the pre-trained C2S Agent without further training. \\
        \bottomrule
    \end{tabular}
\end{table}

We have trained five agent versions depending on the types of available algorithms, as per Table \ref{tab:agents}.

\begin{figure}
    \centering
    \includegraphics[width=\columnwidth]{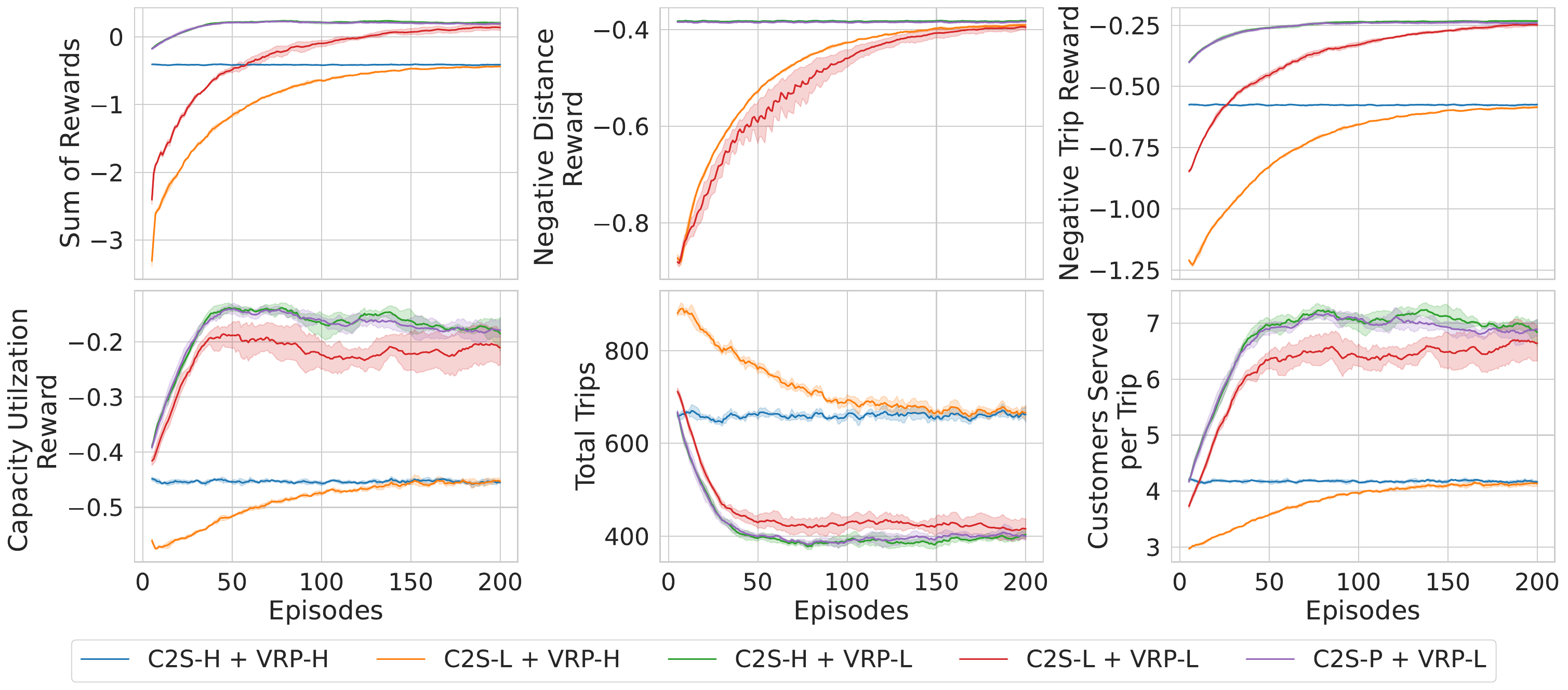}
    \caption{The figure has six plots, divided into two rows. The top row shows the Sum of Rewards, Negative Distance Reward, and Negative Trip Reward from left to right. The bottom row displays Capacity Utilization Reward, Vehicle Trips, and Served Customers per Trip from left to right. For the RL proposed agent, the shaded regions represent the standard deviation across 5 seeds. 
    }
	\label{fig:baselines}
	\vspace{-0.75 em}
\end{figure}


\section{Experiments and Results} \label{sec:results}

\textbf{\textit{Training}}: We trained different agent versions and plotted the training curves to see how they perform across different metrics. The description of each agent is given in Table \ref{tab:agents}. Any versions involving a learning based agent are trained for 200 episodes using an $\epsilon$-greedy approach. Please note that the customer locations are uniformly randomly generated in the 2D grid during training.

Figure \ref{fig:baselines} shows the average training curves for all the different agents, taken over 5 random seeds. For the learning based agents, the shaded regions represent the standard deviation across all random seeds. The top-left figure shows a simple sum of all the reward components. It can be seen that the \textit{C2S-P + VRP-L} and \textit{C2S-H + VRP-L} outperform the rest of the agents and converges pretty fast. Where \textit{C2S-L + VRP-L} converges almost to the similar scale. The other plots provide greater insight into the agent's performance. The top middle-figure plots the straight-line distance reward $D_i$ and is constant for the agents having C2S-H and C2S-P (\textit{C2S-H + VRP-H}, \textit{C2S-H + VRP-L} and \textit{C2S-P + VRP-L}) since for them the C2S agent is not trained. Since the heuristic always picks the nearest warehouse (in the case of C2S-P we make use \textit{C2S-L + VRP-H} which almost learns to pick the customers like the heuristic) this plot shows that the RL agents learn similar warehouse assignment and minimize the distance. But having similar warehouse assignment does not guarantee better customer-vehicle mapping. In the top-right plot, the trip reward $L_i$ depicts that having a learning based VRP agent does better grouping as the trips comprise of more efficient routes by minimizing the overall trip distance. The bottom-left figure depicts the capacity utilization $U_i$, here it is visibly seen that learning-based VRP agents due to their efficient routing are able to utilize the vehicle capacity better compared to VRP-H. The \textit{C2S-H + VRP-L} and \textit{C2S-P + VRP-L} have the best capacity utilization compared to others. The bottom-middle figure displays the total number trips taken by agent for serving all the customers. The learning-based VRP agents outperform the rule-based VRP agent by a margin with almost 40\% lesser trips. All the RL agents initially take more number of trips during the exploration phase due to random warehouse assignments, but it gradually learns to serve the customers in fewer trips with improved and efficient routing. The final bottom-right figure illustrates the number of customers served per trip, here we can see that the agents \textit{C2S-H + VRP-L} and \textit{C2S-P + VRP-L} surpasses the rest of the approaches at the end of training.            

\begin{figure}
    \centering
    \includegraphics[width=0.95\columnwidth]{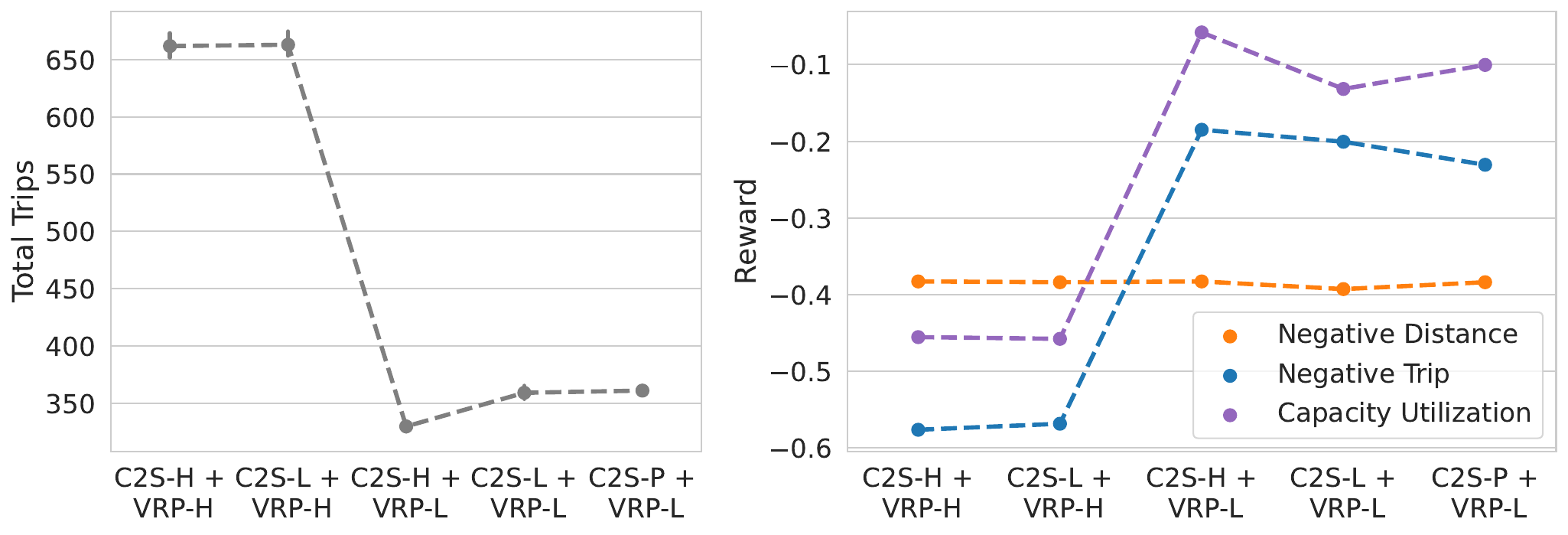}
    \caption{Testing values of metrics averaged over 20 episodes for the different agents. The left figure plots the average number of total trips for different agents, whereas the figure on the right plots the different rewards namely Negative Distance, Negative Trip and Capacity Utilization. }
	\label{fig:testing_metrics}
	\vspace{-1.5em}
\end{figure}

\textbf{\textit{Testing}}: We have also tested the performance of these agents by modifying the customer data distribution by introducing skewness in the probability distribution across the four quadrants. The trained agents are tested for 20 episodes across another set of 3 random seeds. 

In Figure \ref{fig:testing_metrics} we have compared the metrics of the baselines. The figure is split into two, the left figure shows the average number of total trips for each of the agents. We can see here that, the agent \textit{C2S-H + VRP-L} performs the best followed closely by the agents \textit{C2S-L + VRP-L} and \textit{C2S-P + VRP-L}. In the figure on the right, we have plotted together the average of the crucial reward components namely negative distance, negative trip and capacity utilization. Here we can clearly see that the negative distance reward is almost constant for all the five agents. The difference lies in the rest of the two, the \textit{C2S-H + VRP-H} performs the worst with poor negative trip reward and capacity utilization. In the learning-based VRP agents the \textit{C2S-H + VRP-L} outperforms others with better trip reward and vehicle utilization. This is followed by the rest of the two where one component compensate the other. It can stated that having a learning-based VRP agent not only learns better routes but also indirectly benefit the learning-based C2S agent by scheduling efficient routes ultimately leading to better rewards being learnt during exploration. Please note that we do not have any shared information among the two agents we just have reward components derived from the vehicle routing decisions. Additional training and testing results are added in Appendix \ref{subsec:add_results}

To sum it up, the two best performing agents \textit{C2S-H + VRP-L} and \textit{C2S-P + VRP-L} have very similar training and testing results with marginal difference. The only difference between the \textit{C2S-P + VRP-L} agent and \textit{C2S-L + VRP-L} is the way we have trained each of them where the pre-trained agent is trained in two phases (first C2S training followed by VRP training) compared to simultaneous training in the latter. We can infer that training both learning-based agents together reaches a local optima in performance as compared to training each one separately.   

\section{Conclusion and Future Work}

We introduce and compare various methodologies for addressing the combined challenge of fulfillment node selection and vehicle routing within an integrated framework. As far as our knowledge extends, this marks the pioneering attempt to confront this commercially significant issue within the realm of e-commerce, utilizing a structured learning-based approach. Our two learning-based VRP agents consistently outperform the complete heuristic-based approach by a substantial margin. Furthermore, we demonstrate the robustness of these learned models across different customer data distributions, including skewed and random datasets. Given the complexity of this problem, characterized by dynamic demand and a large customer base, opting for a learning-based VRP agent proves to be the most effective strategy, in contrast to simplistic VRP heuristics. Looking ahead, our future research aims to delve deeper into this multi-agent setup, with a specific focus on comprehensive learning-based approaches. We are keen to explore the impact of introducing shared information (features), credit assignment for cooperative training, and the implementation of a central critic shared among the C2S and VRP agents, among other aspects.   


\bibliographystyle{ACM-Reference-Format} 
\bibliography{sample}

\appendix

\section*{Appendix}

This section includes the supplementary material. 

\begin{table}[b]
	\centering
	\begin{tabular}{|l|l|l|}
		\hline
		\textbf{Feature}             & \textbf{Description}                                                                                               & \textbf{Size}         \\ \hline
		
		Embeddings          & Output of GAE & 2 \\ 
		Distance            & Distance $d_{j,i}$ from each warehouse                                          & $N=4$          \\ 
		Local Availability  & Quantity $P_{j}$ of current product available at each warehouse                                                   & $N=4$          \\ 
		Demand              & Customer demand $p_i$ for current product                                                              & 1            \\ 
		
		Time Window &  Interval $[\theta_{i,\min},  \theta_{i,\max}]$  during which order to be fulfilled & 2 \\ 
		
		Clock &  Current timestep in environment & 1 \\ 
		
		Holding Interval & Amount of time the order is put on hold & 1 \\ 
		
		Warehouse Indicator & Vehicle availabaility at depots & $N=4$ \\ \hline
	\end{tabular}%
	\caption{Features in state representation for each customer $c_i$, used by C2S agent as input.}
	\label{tab:c2s_statespace}
\end{table}

\begin{table}[h]
    \centering
    \caption{VRP Agent inputs for evaluating the value of each vehicle-customer pair, with $loc$ as current location, $c$ as proposed customer, and $v$ as the proposed vehicle. All features are normalised as explained in text.}
    \label{tab:vrp_inputs}
    \begin{tabular}{|l|p{11.85cm}|} 
        \hline
        Input & Explanation \\
        \hline
        $d$ & Distance from $loc$ to $c$ \\
        $b_\mathrm{d,short}$ & Is $d$ smaller than neighbourhood radius $\rho$ \\
        $t$ & Time gap from now to start of service at $c$ \\
        $b_\mathrm{t,short}$ & Is time gap within time threshold $\tau$ \\
        $ngb$ & Are $loc$ and $c$ part of the same cluster \\
        $non\_d$ & If $ngb=1$, distance from $c$ to nearest non-member \\
        $c\_left$ & If $ngb=0$, are any in-cluster customers left unserved \\
        $drop_{far}$ & If $c\_left=1$, are the dropped customers farther from depot than $loc$ \\
        $drop_{cls}$ & If $c\_left=1$, are dropped customers within a distance of $\rho$ from $loc$ \\
        $drop_{long}$ & If $c\_left=1$, is the distance from dropped customers to nearest non-member more than distance from $loc$ to dropped customers \\
        $served$ & How many customers of $c$ cluster has $v$ served so far \\
        $cls_{dem}$ & Could $v$ serve all cluster members of $c$ based on demand \\
        $hops$ & How many cluster members of $c$ can be served before $c$ \\
        $cls_{tim}$ & Is every cluster member feasible following $c$ \\
        $urgt$ & How close to time window closure of $c$ is $v$ arriving \\
        $dfrac$ & Ratio of time being consumed for serving $c$ to the fraction of $v$'s capacity being consumed \\
        $remote $& How remote is the neighbourhood of $c$ \\
        \hline
    \end{tabular}
\end{table}

\section{Neural network architecture and training}
\label{subsec:arch}

\textbf{GAE}: The architecture for GAE consists of two graph convoultional layers (2-hop) with the embedding output dimension of 2, with ReLU activation function for hidden layers. The adjacency matrix is constructed by creating positive and negative samples, where positive samples are all instances with edges between two nodes (a relative minority) and negative samples are randomly selected pairs of nodes without edges from the graph (equal in number to the positive samples). The training of the GAE is done iteratively from a buffer of 1000 graphs which are previously generated by running simple heuristic policies for both C2S and VRP functions.

\textbf{C2S-L Agent}: The RL agent is a Deep Q-Learning Network model for customer-warehouse mapping  with 19 inputs mentioned in Table \ref{tab:c2s_statespace} and 5 outputs, as described earlier. The DQN model has a neural network architecture with 4 fully connected layers. The dimensions of the layers are $[19, 76, 38, 5]$, where the first and last layers are the input and output layers respectively. The hidden layers have a \texttt{tanh} activation function, while the output layer has a linear activation function. The model is trained using the \texttt{PyTorch} library in Python 3.8, with an \texttt{Adam} optimizer, a discount factor of 0.9, a batch size of 512, and a mean squared loss function. The experience replay buffer has a capacity of 1,00,000 samples, which are randomly drawn to train the C2S agent. 

\textbf{VRP-L Agent}: The input is a vector of size 17, as specified in Table \ref{tab:vrp_inputs}. This is followed by a fully connected neural network with layers of size (128, 64, 32, 8) and an output layer of size 1 (scalar value). The hidden layers have \texttt{tanh} activation. A memory buffer of maximum 1,00,000 samples is maintained, with training happening after every 4 times every 100 timesteps (basically after completion of each depot routing) with a batch of size 512 samples. The learning rate is 0.001. Exploration is implemented using an $\epsilon-$greedy policy with $\epsilon$ decayed from 1 to 0 with a factor of 0.999 after each episode. Exploration steps are taken uniformly randomly, while exploitation steps are chosen using a \texttt{softmax} function over the values attached to vehicle-customer pairs.

\section{Additional Results}
\label{subsec:add_results}

\begin{figure}
    \centering
    \includegraphics[width=.75\linewidth]{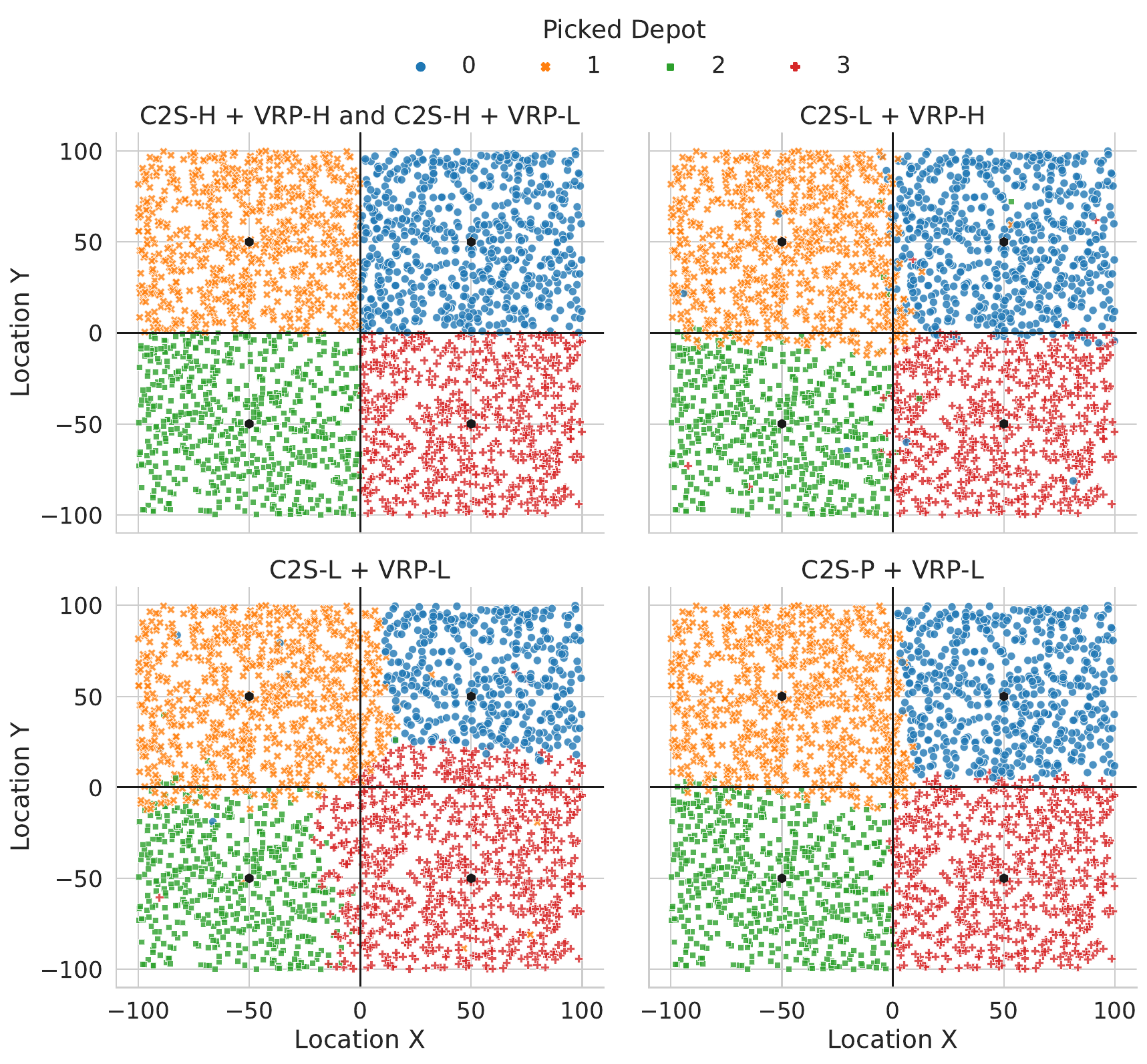}	\label{fig:training_scatter}
    \caption{Scatter plot illustrating the training data with various customer markers, each assigned to a specific depot by the C2S Agent. A black hexagon marks the warehouse location in each quadrant, and bold black lines separate the depot areas for clarity. The figure is divided into four sub-figures, each corresponding to a different baseline mentioned in the subplot title. It's worth noting that this figure represents uniform customer data distribution, with customers equally distributed in all quadrants.}
	\vspace{-1.5em}
\end{figure}
\begin{figure}
    \centering
    \includegraphics[width=.75\linewidth]{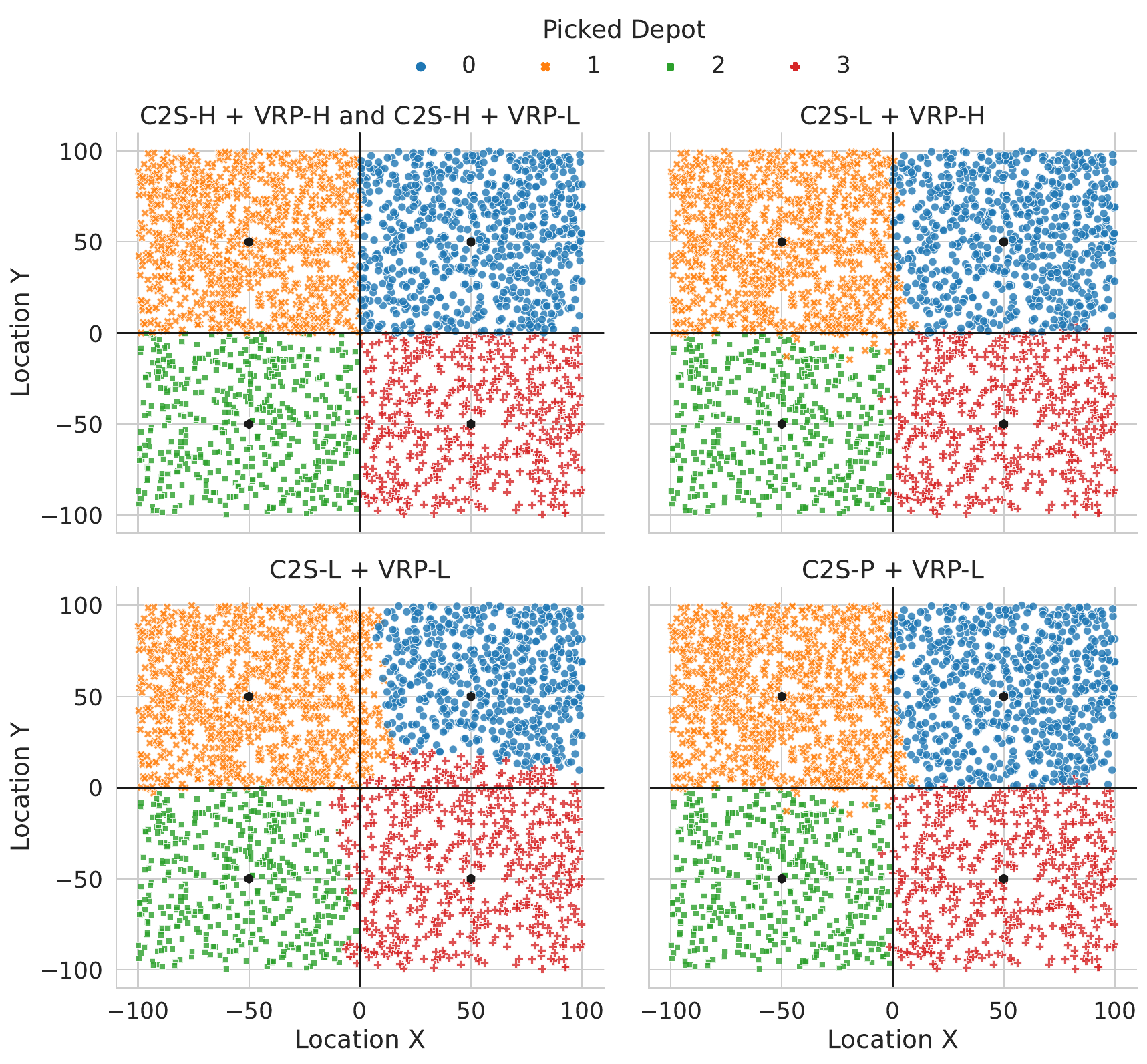}
    \caption{This scatter plot tests various baselines across the entire customer space. Customers are color-coded and marked according to their assigned depot by the C2S Agent. A black hexagon represents the warehouse in each quadrant, and bold black lines separate the depot quadrants for clarity. The figure is divided into four sub-figures, each labeled with its baseline. Note that the customer data distribution is skewed, with more customers concentrated in the upper two quadrants compared to the lower two.}
	\label{fig:testing_scatter}
	\vspace{-1.5em}
\end{figure}

In fig. \ref{fig:training_scatter} and \ref{fig:testing_scatter} we can see that \textit{C2S-H} always assigns the customers to their nearest warehouse, these must be fulfilled by the vehicles originating from that depot as can be seen in the left-top most figure of \textit{C2S-H + VRP-H} and \textit{C2S-H+ VRP-L}. This behaviour of always assigning nearest warehouse is not observed in any \textit{C2S-L agent} baseline, and results show there is an aggregation of the customers of neighbourhood areas. From the scatter we can clearly see this in \textit{C2S-L + VRP-H}, \textit{C2S-L + VRP-L} and \textit{C2S-P + VRP-L}. In some cases, always assigning nearest warehouse may not result in optimal solution and aggregating the customer orders to second nearest warehouse may result in overall less trip distance, utilization of lesser number of vehicles increasing capacity utilization. These observations are true for both the types of customer data distribution (Training - Uniform Data and Testing - Skewed Data).  

\begin{figure}
    \centering
    \includegraphics[width=0.8\columnwidth]{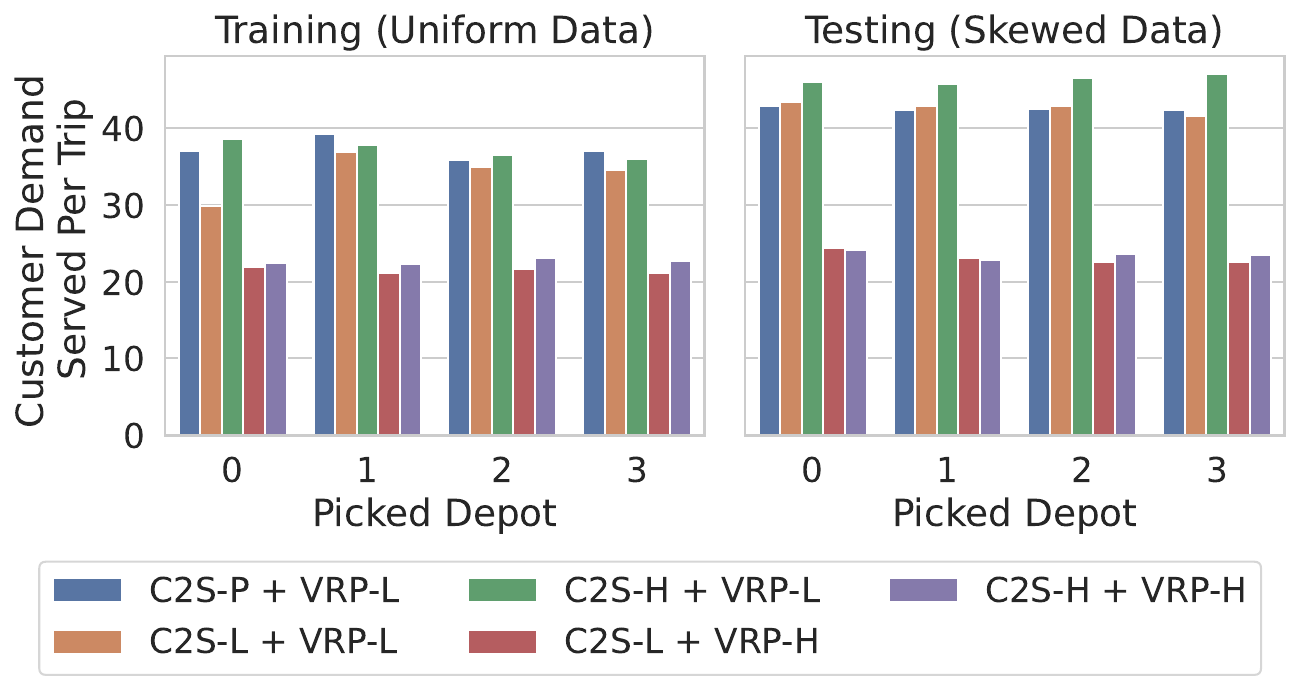}
    \caption{Barplots for showing the average customer demand served per trip for the different agents across depots for both the customer data distributions.}
	\label{fig:demand}
	\vspace{-1.5em}
\end{figure}

Total capacity utilization per vehicle is shown in the figure \ref{fig:demand} Training (Uniform Data), and it can be seen that in the case of agents \textit{C2S-P + VRP-L} and \textit{C2S-H + VRP-L}, total utilization of a vehicle is almost at 80\% followed by \textit{C2S-L + VRP-L} agent. This shows that learning based approach for VRP resulted in better routing of customers, which resulted in utilization at greater capacity. The other two agents \textit{C2S-H + VRP-H} and \textit{C2S-L + VRP-H} resulted in the usage of only 40\% of vehicle capacity, which clearly indicates that incorporating \textit{VRP-L} resulted in the increase of vehicle capacity utilization by almost double. The similar behaviour is observed in Testing(Skewed Data), where the capacity utilization per vehicle at each varies by large margin in case of agents with learning agents compared with the heuristic approaches.

\end{document}